\documentclass[a4paper]{svproc}
\usepackage{hyperref}
\hypersetup{
    colorlinks=true,
    linkcolor=blue,
    filecolor=magenta,      
    urlcolor=cyan,
    pdftitle={Overleaf Example},
    pdfpagemode=FullScreen,
}
\usepackage{url}

\usepackage{times}
\usepackage{epsfig}
\usepackage{graphicx}
\usepackage{amsmath}
\usepackage{amssymb}
\usepackage{algorithm}
\usepackage{algpseudocode}
\usepackage{graphics}
\usepackage{verbatim}
\usepackage{xcolor}
\usepackage{caption}
\usepackage{subcaption}
\usepackage{cite}
\newcommand{\argminE}{\mathop{\mathrm{argmin}}}
\DeclareRobustCommand*{\ora}{\overrightarrow}

\newenvironment{myitem}{\begin{list}{$\bullet$}
{\setlength{\itemsep}{0pt}
\setlength{\topsep}{0pt}
\setlength{\leftmargin}{12pt}
\setlength{\parsep}{0pt}
\setlength{\itemsep}{0pt}
\setlength{\partopsep}{0pt}}}%
{\end{list}}

\begin{document}
\mainmatter              
\title{6N-DoF Pose Tracking for Tensegrity Robots}
\titlerunning{Tensegrity Robot Perception}  
%
\author{Shiyang Lu\inst{1}, William R. Johnson III\inst{2}, Kun Wang\inst{1}, Xiaonan Huang\inst{2},\\
Joran Booth\inst{2}, Rebecca Kramer-Bottiglio\inst{2}, Kostas Bekris\inst{1}}
\authorrunning{Shiyang Lu et al.} 
%
\tocauthor{Shiyang Lu\inst{1}, William R. Johnson III\inst{2}, Kun Wang\inst{1}, Xiaonan Huang\inst{2},
Joran Booth\inst{2}, Rebecca Kramer-Bottiglio\inst{2}, Kostas Bekris\inst{1}}
\institute{Rutgers University, New Brunswick, NJ, USA,
\footnote{\{shiyang.lu, kun.wang2012, kostas.bekris\}@rutgers.edu}
\and
Yale University, New Haven, CT, USA
\footnote{\{will.johnson, xiaonan.huang, joran.booth, rebecca.kramer\}@yale.edu}}

\maketitle              

\vspace{-.2in}
\begin{abstract}
Tensegrity robots, which are composed of compressive elements (rods) and flexible tensile elements (e.g., cables), have a variety of advantages, including flexibility, low weight, and resistance to mechanical impact. Nevertheless, the hybrid soft-rigid nature of these robots also complicates the ability to localize and track their state. This work aims to address what has been recognized as a grand challenge in this domain, i.e., the state estimation of tensegrity robots through a marker-less, vision-based method, as well as novel, on-board sensors that can measure the length of the robot's cables. In particular, an iterative optimization process is proposed to track the 6-DoF pose of each rigid element of a tensegrity robot from an RGB-D video as well as endcap distance measurements from the cable sensors. To ensure that the pose estimates of rigid elements are physically feasible, i.e., they are not resulting in collisions between rods or with the environment, physical constraints are introduced during the optimization. Real-world experiments are performed with a 3-bar tensegrity robot, which performs locomotion gaits. Given ground truth data from a motion capture system, the proposed method achieves less than 1~cm translation error and 3 degrees rotation error, which significantly outperforms alternatives. At the same time, the approach can provide accurate pose estimation throughout the robot's motion, while motion capture often fails due to occlusions. 
\vspace{-.1in}
\end{abstract}

\begin{figure}[t]
    \centering
    \includegraphics[width=\textwidth]{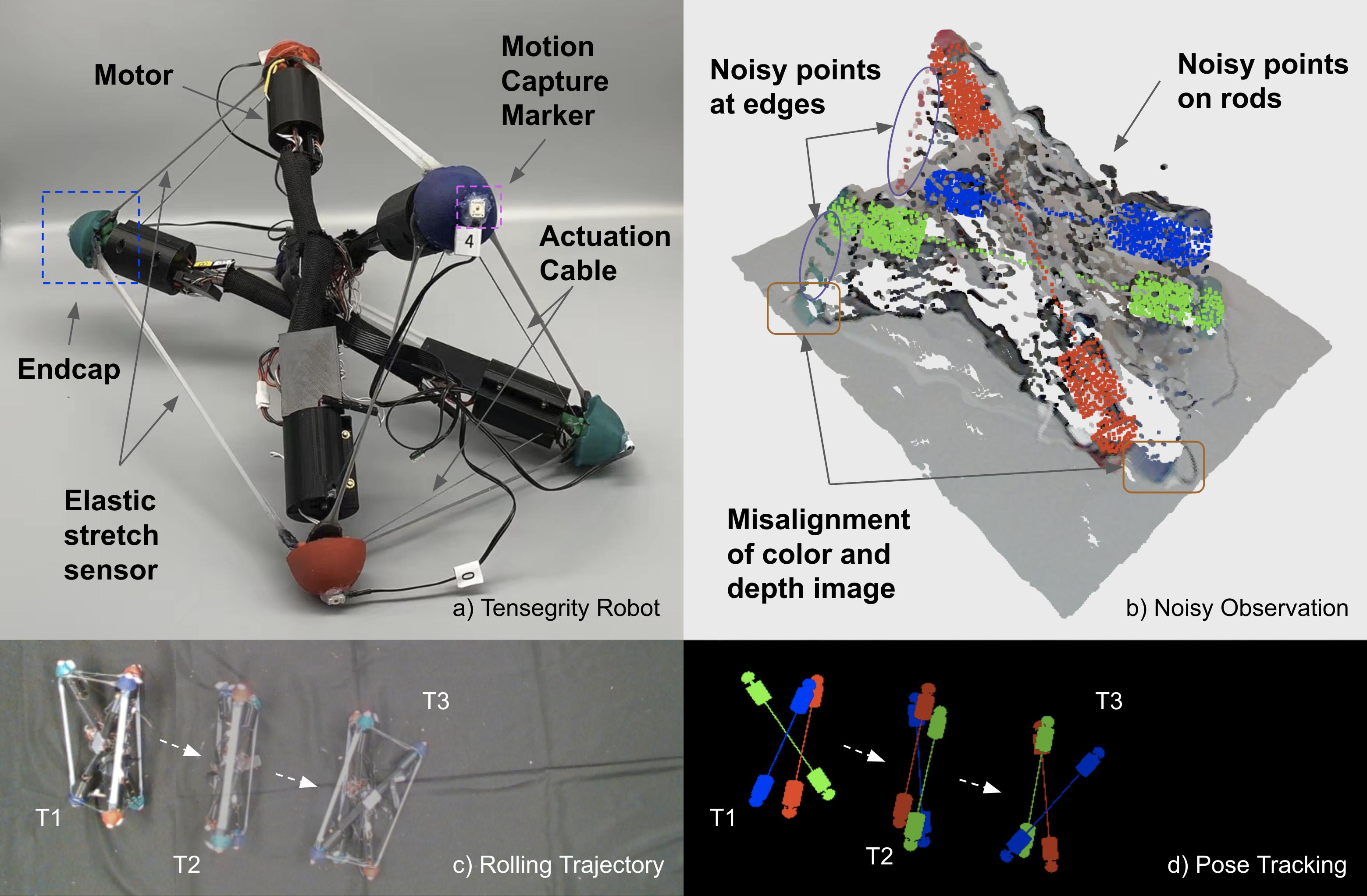}
    \vspace{-.25in}
    \caption{\footnotesize \textbf{(a)} The 3-bar tensegrity robot used in the accompanying experiments. Each bar has a length of 36~cm. The robot can reshape itself and perform locomotion by changing the length of its actuation cables. Distances between endcaps are measured by on-board elastic stretch sensors. Motion capture markers are used to generate ground truth poses only for the purpose of evaluation. \textbf{(b)} Noisy observations from a commodity RGB-D camera pose challenges to vision-based pose estimation. \textbf{(c)} The tensegrity robot rolling on a carpet. The shape of this robot keeps changing during the motion and generates significant self-occlusions. \textbf{(d)} 6-DoF pose estimates for each bar at the corresponding time steps.}
    \label{fig:intro}
    \vspace{-.3in}
\end{figure}

\vspace{-.4in}
\section{Introduction}\vspace{-.1in}
Tensegrity robots are lightweight, deformable platforms that are composed of compressive elements (rods) and flexible tensile elements (e.g., cables~\cite{vespignani2018design}, springs~\cite{kim2017design}, elastic membranes~\cite{baines2020rolling}, etc.). This enables tensegrity robots to distribute external forces through deformation and thus prevent damage. Furthermore, by controlling its tensile elements, a tensegrity robot can reshape itself and move in complex environments~\cite{vespignani2018design}. An illustration of a 3-bar tensegrity robot, used for the experiments presented in this work,  is shown in Fig.~\ref{fig:intro}. Tensegrity robots' adaptability and compliance have motivated their use in many applications, such as manipulation~\cite{lessard2016bio}, locomotion~\cite{sabelhaus2018design}, morphing airfoil~\cite{chen2020design} and spacecraft landers~\cite{bruce2014superball}. While their versatility is exciting, tensegrity robots are difficult to track, model and control because of their complex dynamics~\cite{shah2021tensegrity}.

In particular, state estimation and tracking are critical for control algorithms that require the current robot state. For tasks, such as bringing the tensegrity to a desired goal configuration, it is necessary to identify the 6D pose of its rigid elements. Reconstructing the trajectory of a tensegrity robot can also be used for sim2real, where the parameters of a simulated model are identified to match real-world trajectories. Such simulated models~\cite{NTRTSim,pmlr-v120-wang20b, wang2021sim2sim, wang2022recurrent} can then be used to generate locomotion policies via RL~\cite{surovid2018any, surovik2019adaptive} without the need for many real world experiments. 

Accurately estimating the state of a tensegrity robot, i.e., the 6-DoF pose of each rod, remains a grand challenge in this domain~\cite{shah2021tensegrity}. Vision-based pose estimation is especially challenging due to noisy observations, self-occlusions, and a large number of degrees of freedom as shown in Fig.~\ref{fig:intro}. This drives the need for improved sensor fusion, physics-aware state estimation algorithms, and more stable, on-board sensors~\cite{shah2021tensegrity}. 

This work proposes a robust approach for markerless, vision-based pose tracking of an $N$-bar tensegrity robot given as input: 1) RGB-D images from a single, external commodity camera, and 2) measured endcap distances from recently developed, on-board, elastic stretch sensors.  The proposed approach has the following key features:  
\begin{myitem}
    \item An ICP-like tracking method is used to fuse the multi-modal inputs in an iterative optimization process. An objective function is designed to maximize both unary and binary potentials. Unary potentials reflect how well the mesh models of the rods' endcaps fit with the observed point cloud. Binary potentials reflect the fitness of predicted endcap distances and the measured ones from on-board sensors. Adaptive weights are designed to balance these two potentials.
    \item A variety of physical constraints between rigid elements of tensegrity robots are introduced in the optimization. These constraints ensure that the estimated poses are physically feasible without relying on a sophisticated physics engine and further improve pose estimation accuracy.
\end{myitem}
The proposed method is evaluated qualitatively and quantitatively on real-world experiments discussed in section~\ref{experiments}, using a 3-bar tensegrity performing locomotion gaits. The method achieves low translation and rotation error when compared against alternatives as well as ground truth data from motion capture. More critically, the approach can provide pose estimates throughout the robot's motion, while motion capture often fails due to occlusions. A \href{https://sites.google.com/view/tensegrity-robot-perception}{link}\footnote{\href{https://sites.google.com/view/tensegrity-robot-perception}{https://sites.google.com/view/tensegrity-robot-perception}} points to a website that highlights the available experiments and shares an open-source dataset with the corresponding trajectories. The hope is that the methodology will prove useful for additional platforms beyond tensegrities that can make use of on-board distance sensors for soft elements and an external RGB-D camera.


\vspace{-.1in}
\section{Related Work}\label{related_work}\vspace{-.1in}
The majority of state estimation work on tensegrity robots is based on multiple on-board sensors, such as sensors for contact sensing~\cite{caluwaerts2014design}, cable length measurement~\cite{tietz2013tetraspine}, strut extension~\cite{friesen2014ductt}, and inertial measurement units (IMUs)~\cite{caluwaerts2014design, bruce2014design}. However, with on-board sensors only, it is hard to reconstruct the state of tensegrity robots in the global frame. To estimate states in the global frame, Caluwaerts et.al~\cite{caluwaerts2016state} proposed a variant of the Unscented Kalman Filter (UKF) algorithm by fusing the measurements from IMU sensors and ranging sensors that are placed at the endcaps of rods and fixed anchors (landmarks) at the testing area. Similarly, a motion capture system could be setup and track the markers placed on endcaps of rods~\cite{booth2021surface} to get their global coordinates. Despite being accurate when all the markers are visible, tracking often fails when markers are occluded. Moreover, since a motion capture system is fixed once calibrated, it constrains the movement of a robot to a small region.

More recently, Moldagalieva et al.~\cite{8700452} proposed the first vision-based method to estimate the pose of a tensegrity robot using fiducial markers. In particular, a hemispherical camera module was attached to the base of their robot, so as to track the fiducial markers printed on a triangle-shaped plate fixed at the upper part of the robot. The pose estimates could be further fine-tuned by fusing the cable length measurement using a neural network~\cite{8932581}. However, the requirement for fiducial markers to always be visible to the camera prevents the robot from doing actions like rolling. It also requires a large amount of data for training a deep model. In contrast, the method proposed in this paper detects endcaps of rods based on color images instead of relying on fiducial markers.

ICP algorithms~\cite{besl1992method, chen1992object} have been widely used in the industry for point cloud registration and model matching. A large number of variants~\cite{rusinkiewicz2001efficient, segal2009generalized, park2017colored} have been proposed since the advent of this algorithm for better accuracy and faster convergence. Some extensions of the ICP algorithm can even be applied to track articulated objects, such as human bodies~\cite{mundermann2007accurately} and hands~\cite{tagliasacchi2015robust}. This proposed work is in line with the direction of those methods but with an objective function and physical constraints that are specifically designed for tensegrity robots.

\vspace{-.1in}
\section{Problem Formulation}\label{problem_formulation}\vspace{-.1in}


This work aims to track the pose of an $N$-bar tensegrity robot, i.e. 6-DoF pose of each rigid element (rod) of that robot, which are connected with flexible tensile elements (cables). An illustration of the robot used for the experimental tests of this paper is shown in Fig.~\ref{fig:intro}. Two types of sensors are considered: 1. An overhead RGB-D camera records the motion of the tensegrity robot; 2. on-board strain sensors are publishing the cable length measurements during the motion. The images and cable length measurements are synchronized. The mesh model $\mathcal{M}$ of the rods and endcaps is assumed available, and the structure of the robot (i.e., the connectivity between endcaps with cables) is also provided. The endcaps of different rods are painted with different colors, and the ranges of their HSV values are known. Then, the objective is to compute the set of poses $P_t = \{p_t^1, ..., p_t^N\} \in SE(3)$ of each rod at the current time step $t$, given:
\begin{myitem}
    \item known parameters for the setup, i.e., the camera's intrinsic matrix $C$, the tensegrity's rod length $l_{rod}$ and diameter $d_{rod}$,
    \item previous rod pose estimates at step $t - 1$, i.e., $P_{t-1} = \{p^1_{t-1}, ..., p^N_{t-1}\} \in SE(3)$, 
    \item current measurements at time step $t$:
    \begin{itemize}
        \item the input of an RGB-D image $I_t$ observing the tensegrity, 
        \item cable length measurements $L_t = \{l_t^{ij} | (i, j) \in E\}$ where $i,j$ are endcap indices and $(i, j) \in E$, if endcap $i$ and endcap $j$ are connected by a cable.
    \end{itemize}
\end{myitem} 
To initialize the process, a 2D Region of Interest (RoI), in the form of a bounding box, is manually defined for each endcap  at the initial frame $I_0$ of the video sequence.

\vspace{-.1in}
\section{Method and Implementation}\label{method}\vspace{-.1in}

The following discussion describes step by step how the poses $P_t$ are estimated at each time step. In terms of the visual data and given that the depth measurements of rods can often be very noisy, as shown in Fig.~\ref{fig:intro}b, only the observations of endcaps are used in the proposed state estimation process.

\vspace{-.2in}
\subsection{Pose Initialization}\label{initialization}
\vspace{-.05in}
The proposed tracking algorithm first establishes an initial pose estimate $P_{init}$ given the manually defined image RoIs. This initial estimate does not have to be very accurate since it will be optimized by consecutive steps of the iterative process. The foreground pixels in each RoI are segmented based on the range of HSV values for that endcap, and then projected to 3D points given the camera intrinsic matrix $C$ and the depth information. The centroid of the observed point cloud of each endcap is computed and used as a rough estimate of the initial endcap position. The initial position of each rod is then estimated by simply using the center of its two endcaps. The initial rotation is estimated using axis-angle representation where the axis is aligned with the direction from one endcap to the other and the angle is arbitrarily set due to rotational symmetry.

\vspace{-.2in}
\subsection{Iterative Pose Tracking}
\vspace{-.05in}
The proposed method addresses the pose tracking problem through an iterative optimization process where each iteration is composed of a transition step and a correction step. Those two steps are performed alternately until convergence or a preset maximum number of iterations is reached (an empirical number is 6). In the transition step, only the RGB-D observation is used to estimate a local transformation from the previous time step. While in the correction step, a joint constrained optimization is performed given the input of estimated endcap positions and cable length measurements. Those two steps are detailed below. 

\vspace{.05in}
\noindent \textbf{1) Transition Step}\\
The transition model, also known as the action model, aims to provide a 6-DoF local transformation $\delta P_t^i$ for rods $i=1,2,..,N$, such that the new pose estimate for each rod to be $\hat{P}_t^i = \delta P_t^i \cdot P_{t-1}^i$. It is, however, not easy to compute this transformation directly from control gaits, due to the complex dynamics of tensegrity robots. Instead, the local transformation is estimated via scan matching, i.e., the registration of endcap model points with the observed points in the RGB-D data. To prevent unobserved model points (e.g., those on the backside of an endcap) from being registered with the point cloud, this work adopts a hidden point removal method~\cite{hidden_point_removal} so that only visible points on the model will be used for registration and thus improve accuracy.



Correspondence between endcap model points and observed points is established based on nearest neighbors as long as they are within a maximum distance threshold in the 3D space and also within an HSV range in the color space. The maximum distance threshold $d_{max}$ is designed to be a dynamic value that decreases with the number of iterations. The intuition behind this is that in the first few iterations the region for correspondence search should be large in case of significant movement, while in later iterations it should be small for fine-grained estimates. With known correspondences, a local transformation for each rod can be estimated with the Kabsch algorithm~\cite{kabsch1976solution}, which minimizes the weighted root mean square error (RMSE) between point pairs. The weight of each point pair is computed using the following equation, where $d$ is the distance between a pair of points. This weight is monotonically decreasing when the distance increases and clipped to zero if the distance goes beyond $d_{max}$.
\vspace{-.1in}
\begin{equation}
    w(d) = \left\{
    \begin{aligned}
    1 - (\frac{d}{d_{max}})^2, & & d \leq d_{max} \\
    0, & & d > d_{max}
    \end{aligned}
    \right.
\end{equation}
\vspace{-.15in}

It is often the case, however, that the corresponding points of one endcap are too few for registration, mainly due to occlusions. To handle this problem, a certain number of dummy points (an empirical number is 50) are randomly sampled from observed and model points at the previous time step and pasted to the points in the current time step. This step is effective in preventing arbitrary poses being generated and results in more stable state estimates. It is also arguably better than alternative motion models, such as assuming the rod is not moving at the current time step since the observation of the other endcap can provide meaningful information. The weights of those dummy points are half the value of observed points.

\vspace{.05in}
\noindent
\textbf{2) Correction Step}\\
In the correction step, a joint constrained optimization is performed. The intuition behind this optimization is that the estimated endcap positions are expected to be close to the output of the transition model, and the distance between two connected endcaps should be close to the cable length measurement. Meanwhile, endcaps and rods should satisfy a variety of physical constraints to ensure that the estimated 6-DoF poses are valid. Given the current pose estimates $\hat{P}_t = \{\hat{p}^1_{t}, ..., \hat{p}^N_{t}\}$ from the transition step and cable length measurements $L_t$, the objective function shown below is aiming to find a set of optimal endcap positions $Q = \{q^1_t, ..., q^{2N}_t\}$:
\vspace{-.1in}
\begin{equation}
    \begin{gathered}
        L(Q) = \sum_i^{2N}w_i||q^i_t - \hat{q}^i_t||^2_2 + \sum_{(i, j) \in E}w_{ij}(||q^i_t - q^j_t|| - l^{ij}_t)^2\\
        Q_{optim} = \argminE_{Q}L(Q) \vspace{-.1in}
    \end{gathered}
\end{equation}
where $\hat{Q} = \{\hat{q}^1_t, ..., \hat{q}^{2N}_t\}$ are estimated endcap positions from the transition model and $w_i$, $w_{ij}$ are adaptive weights for the unary and binary loss terms respectively. The constraints that are accompanying this objective function and computation of those dynamic weights are elaborated upon in subsections \ref{physical_constraints} and \ref{adaptive_weights} respectively.

Once the positions of endcaps are optimized, the current pose estimates are updated for the next iteration such that the position of the rod is set to be the center of its endcaps and the rotation axis is set to be aligned with the direction from one endcap the other. Different from the initialization step in subsection \ref{initialization}, the rotation angle is no longer set to be arbitrary but rather to a value that minimizes the geodesic distance in $SO(3)$ from the previously estimated rotation. Despite rotational symmetry, an arbitrary rotation angle can affect downstream tasks that need to compute angular velocity, such as feedback control and system identification. This constraint optimization is performed with sequential least squares programming (SLSQP).

\vspace{-.1in}
\subsection{Physical Constraints}\label{physical_constraints}
\textbf{1) Constraints between endcaps}\\
Rods are rigid elements of tensegrity robots. The distance between the estimated endcap positions of a rod should be a constant, i.e., the rod length $l_{rod}$, regardless of time. Assuming that $i, j$ are the two endcaps of a rod, an equality constraint is applied: \vspace{-.1in}
$$
||q_i - q_j|| = l_{rod} \vspace{-.1in}
$$

\noindent
\textbf{2) Constraints between rods}\\
The estimated poses of rods should not make them collide or even penetrate each other between consecutive frames. Thanks to the simple geometry of the rods, the physical constraints between them can be added to the optimization problem without using a sophisticated simulation engine.

Each rod is approximated as a cylinder and its endcaps as two hemispheres. With this approximation, a rod can be parameterized with only two variables: its length $l_{rod}$ and its diameter $d_{rod}$. A collision between rod $i$ and rod $j$ can be easily detected by comparing the rod diameter $d_{rod}$ with the rod distance $d(c_t^i, c_t^j)$ where $(c_t^i, c_t^j)$ is the closest pair of points on the main axis of rod $i$ and rod $j$ at time step $t$. Rod $i$ and rod $j$ are colliding with each other if $d(c_t^i, c_t^j) < d_{rod}$. The closest pair of points can be computed given the rod pose estimates at timestep $t$. An inequality constraint can be added, i.e.
\vspace{-.2in}
\begin{align*}
|\ora {c_t^ic_t^j}| \geq d_{rod} \tag{a} \vspace{-.1in}
\end{align*}
Meanwhile, the relative poses between two rods are not expected to change dramatically, and it is impossible for one rod to suddenly cross through another between two consecutive frames. An illustration of penetration between two rods is shown in Fig.~\ref{fig:rod_constraints}. Such penetration can be detected if the angle between $\ora {c_t^ic_t^j}$ and $\ora {c_{t-1}^ic_{t-1}^j}$ is greater than 90 degrees, which can be written as $\ora {c_t^ic_t^j} \cdot \ora {c_{t-1}^ic_{t-1}^j} < 0$. Thus, another inequality constraint can be added, i.e.,
\vspace{-.2in}
\begin{align*}
\ora {c_t^ic_t^j} \cdot \ora {c_{t-1}^ic_{t-1}^j} \geq 0 \tag{b}
\end{align*}
\begin{figure}[H]
    \vspace{-0.5in}
    \centering
    \includegraphics[width=0.6\textwidth]{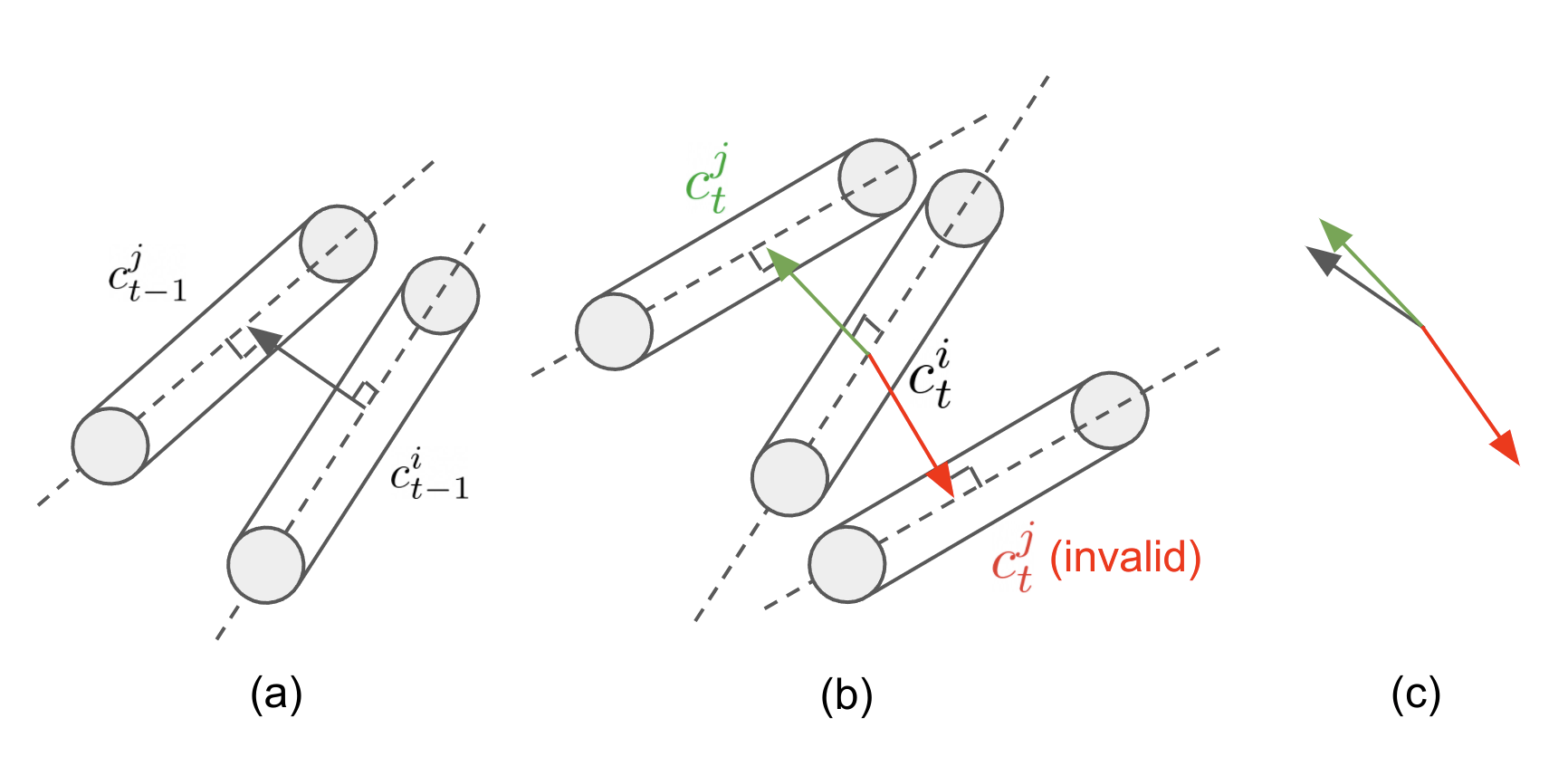}
    \vspace{-.2in}
    \caption{\footnotesize \textbf{Illustration of the physical constraints between rods.} \textbf{(a)} Rod $i$ and rod $j$ at time step $t - 1$, where the $(c_{t-1}^i, c_{t-1}^j)$ is the closest pair of points. \textbf{(b)} Rod $i$ and rod $j$ with two pose hypotheses at time step $t$. Their distance should satisfy the constraint $|\ora {c_t^ic_t^j}| \geq d_{rod}$. \textbf{(c)} Despite that both hypotheses for rod j satisfy the distance constraint, the one in red is physically infeasible as the angle between $\ora{c_t^ic_t^j}$ and $\ora{c_{t-1}^ic_{t-1}^j}$ is greater than 90 degrees, which means rod $j$ crosses through rod $i$ between two consecutive frames.}
    \vspace{-.25in}
    \label{fig:rod_constraints}
\end{figure}
Nevertheless, it is difficulty to directly apply constraints $(a)$ and $(b)$ despite their simple mathematical formulas. This is because the closest pair of points $(c_t^i, c_t^j)$ depends on the relative pose of two rods, and their positions are non-linear with respect to the endcap positions. This problem is addressed by introducing a linear approximation of the constraints based on the assumption that the change in the closest points on the two rods is negligible in the rod's local frame between two consecutive frames, and the angle $\theta$ between $\ora {c_t^ic_t^j}$ and $\ora {c_{t-1}^ic_{t-1}^j}$ is small, i.e. $cos(\theta) \approx 1$. With this assumption, constraints $(a)$ and $(b)$ can be approximated with a single constraint linear to the current endcap positions $(c)$, i.e., \vspace{-.2in}
\begin{align*}
\ora {\hat{c}_t^i\hat{c}_t^j} \cdot \frac{\ora {c_{t-1}^ic_{t-1}^j}}{ |\ora {c_{t-1}^ic_{t-1}^j}|} \geq d_{rod} \tag{c}
\end{align*}
where $\hat{c}_t^i$ and $\hat{c}_t^j$ are the same points as $c_{t-1}^i$ and $c_{t-1}^j$ in the local frame of rod $i$ and rod $j$. This linear approximation makes it easier to optimize, and it is shown to be effective in experiments. If $(c_{t-1}^i, c_{t-1}^j)$ are the endpoints of rod $i$ and rod $j$, then this inequality constraint is not applied. 

\vspace{.1in}
\noindent
\textbf{3) Constraints between rods and the supporting plane}\\
If given the prior knowledge that the tensegrity robot is not moving in a complex environment but instead on the ground, an additional physical constraint can be applied between the rods and the ground to prevent invalid poses that make rods penetrate the ground. A extrinsic transformation from the camera frame to the world frame, which is composed of a rotation $\bar{R}$ and a translation $\bar{t}$, can be computed using plane detection by RANSAC, where the ground is represented as $z = 0$ in the world frame. The inequality constraint for endcaps at position $q^i_t$ is: \vspace{-.05in}
$$
z(\bar{R}q_t^i + \bar{t}) \geq d_{rod} / 2, \hspace{.1in} i = 1,2,..,2N
\vspace{-.25in}$$

\noindent \subsection{Adaptive Weights}\label{adaptive_weights}\vspace{-.05in}
Adaptive weights are applied to balance the unary loss and binary loss in the objective function in the correction step. Let the ratio $r_i$ for the endcap $i$ be the number of corresponded points over the total number of rendered points for endcap $i$ assuming no occlusion. The unary weight $w_i$ and binary weights $w_{ij}$ are computed as follows:
\vspace{-.15in}
$$
w_{i} = max(r_i, \hspace{0.05in} 0.1)
$$
\begin{equation*}
    w_{ij} = \left\{
    \begin{aligned}
    0, & & r_i > 0.5 \hspace{0.05in} and \hspace{0.05in} r_j > 0.5 \\
    0.25, & & r_i < 0.2 \hspace{0.05in} or \hspace{0.05in} r_j < 0.2 \\
    0.25 * (1 - 0.5(r_i + r_j)), & & otherwise
    \end{aligned}
    \right.
\end{equation*}

The design logic is that endcap distance measurements by on-board elastic stretch sensors are often not accurate enough and may enlarge error when both the end caps of a rod are visible, despite that these measurements are rather helpful when endcaps are occluded. When the number of corresponding points for both endcaps of a rod is large, i.e. $r_i > 0.5$ and $r_j > 0.5$, point registration is likely to give an accurate estimate, and the loss introduced by the distance measurement is not considered. However, when either of the endcaps on a rod is heavily occluded, i.e. $r_i < 0.2$ or $r_j < 0.2$, then the distance measurement must be taken into account and is set to be a constant value (0.25). The binary weights are otherwise linear to the sum of the ratio $r_i$. These coefficients and thresholds are empirical, but they work sufficiently well in the experiments.

\vspace{-.2in}
\subsection{Noise Filtering}\label{noise_filtering}
\vspace{-.1in}
Sensor noise in modern Lidar-based RGB-D cameras is typically small. Nevertheless, it is observed in real experiments that misalignment of the color and depth images does occur, and the depth quality is bad at the object edges. An example is shown in Fig.~\ref{fig:intro}(b). This noise is usually not a big issue for target objects with a large observable region, but it turns out to be rather problematic for tensegrity robots as the endcaps are quite small (radius = 1.75~cm in the experiment). Although this noise can be partially removed by careful tuning of the camera setup, it is sensitive to many uncontrollable factors, such as lighting conditions in real environments. To minimize the sensor noise, observed points for each endcap are filtered given the prior knowledge of the endcap size. To be more specific, observed points are first filtered by the HSV range in color space and max correspondence distance in 3D space. Noisy points are further removed if their depths are greater than the minimum depth of the remaining points plus the endcap radius.


\vspace{-.2in}
\section{Experiments}\label{experiments}
\vspace{-.1in}
\subsection{Hardware Setup}
\begin{figure}[h]
    \centering
    \includegraphics[width=0.6\textwidth]{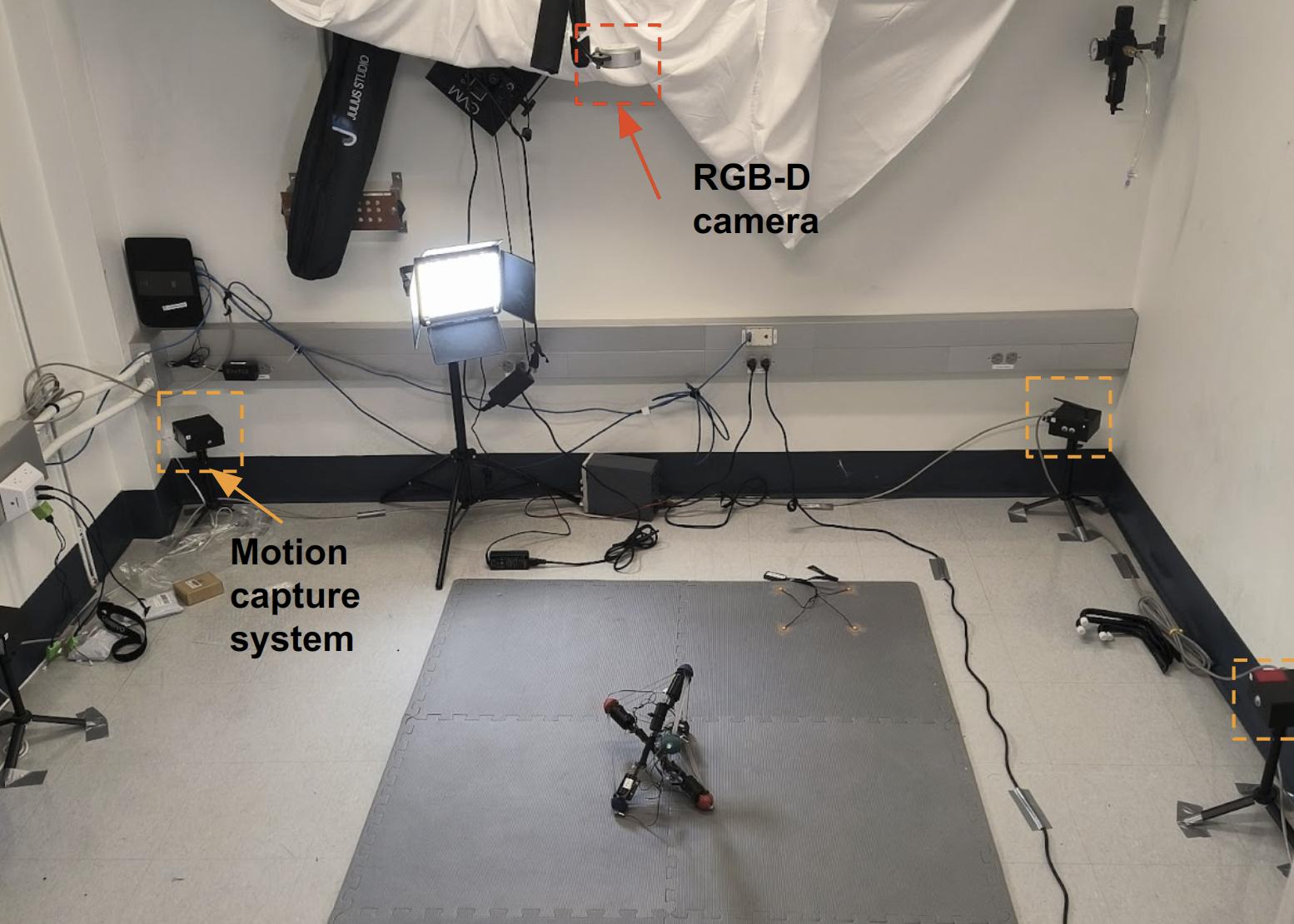}
\vspace{-.1in}
\caption{\footnotesize \textbf{Hardware setup of the experiments.} An RGB-D camera (RealSense L515) is placed overhead to capture the whole region where the 3-bar tensegrity robot performs locomotion. A motion capture system (Phasespace, Inc.) with six cameras is placed around this region to capture endcap markers and generate ground truth for the purpose of evaluation.}
    \label{fig:hardware_setup}
    \vspace{-.4in}
\end{figure}

\vspace{-.05in}
The real-world data are collected on a 3-bar tensegrity robot with nine elastic stretch sensors connecting the nodes. The length of of each bar is 36 cm. Six of the sensors run parallel to the six cables that actuate the robot while the other three longer sensors are passive.  The sensors are made from silicone elastomers (Eco-Flex 30 or DragonSkin 20; Smooth-On, Inc.) encapsulating two layers of a liquid metal paste made from eutectic gallium indium.  The two layers of liquid metal compose a parallel plate capacitor whose capacitance varies linearly as the sensor is stretched~\cite{johnson2021integrated}.  The capacitance of each sensor is measured on-board with a commercial capacitive sensing breakout board (MPR121; Adafruit).  The sensors are calibrated individually to obtain a linear model mapping the sensor's capacitance to its length.  Then, the sensors are used for feedback control during actuation trajectories, and the length of each sensor is published at each time step.  The color and depth image sequences are collected by an RGB-D camera (Intel Realsense L515) at 720p resolution. In order to get the ground truth poses for quantitative evaluation, a commercial motion capture system (Phasespace, Inc.) is used. Markers are placed on the endcaps of rods to measure their positions. Note that measurements from the motion capture system also contain some small error, but this error is difficult to quantify. Fig.~\ref{fig:hardware_setup} shows the hardware setup and the environment where the robot experiments are performed.

\vspace{-.25in}

\vspace{-.1in}
\subsection{Data Collection}\label{dataset}
\begin{figure}[t]
    \centering
    \includegraphics[width=0.9\textwidth]{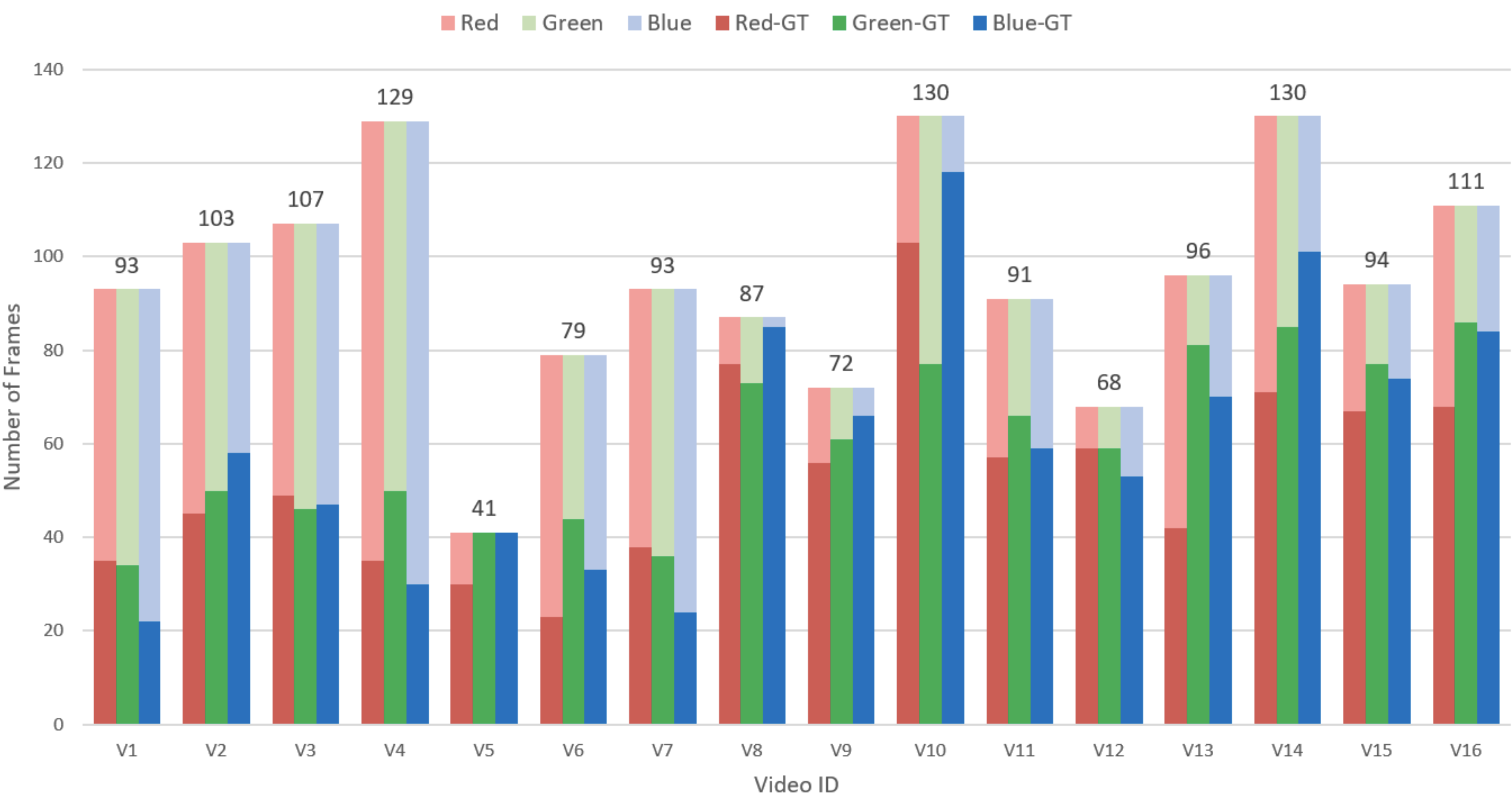}
        \vspace{-.1in}
    \caption{\footnotesize \textbf{Statistics of the dataset.} The dataset contains 16 videos that are captured around 10Hz. This plot shows the number of frames in each video, and the number of available ground truth poses for rods with R/G/B endcaps respectively. The ground truth pose of a rod is only available when both of its endcap markers are visible to the motion capture system, which is not often the case in real experiments. This, however, motivates the proposed work.}
    \label{fig:gt_statistics}
    \vspace{-.3in}
\end{figure}

\vspace{-.1in}
The proposed algorithm is evaluated on a set of self-collected robot trajectories. In total, 16 unique trajectories are collected where the tensegrity robot is placed at different starting positions and a different set of controls are performed. For each trajectory, a sequence of RGB-D images and their corresponding cable length measurements are aligned and recorded at a frame rate around 10Hz. The shortest sequence contains 41 frames while the longest contains 130 frames. In total, this dataset contains 1524 frames, and about 95 frames on average for each trajectory. Note that markers placed on endcaps are not always visible to the motion capture system. The visibility of a marker from the camera and from the motion capture system is, however, different. An endcap occluded from the camera may have its marker visible to the motion capture system and vice versa. In total, the number of ground truth poses for all rods in all videos is 2786. Detailed statistics of the dataset are shown in Fig.~\ref{fig:gt_statistics}.

\vspace{-0.2in}
\subsection{Evaluation Metric}\vspace{-.1in}
For quantitative evaluation of pose estimation, the mean and standard deviation of both translation and rotation errors are reported. Due to rotational symmetry along the main axis of rods, the rotation error is evaluated using the angle between the estimated and ground truth main axis, i.e. the pose errors of rods are evaluated on five degrees of freedom. Pose estimation results are also evaluated using the ``2cm-5deg'' metric, which reports the percentage of pose estimates that are within both 2cm translation error and 5 degree rotation error. The translation error of the center of mass is also evaluated, where the center of mass of the robot is computed by averaging the center positions of all rods. Apart from the pose estimation result in the global frame, the local shape reconstruction result is also evaluated. This result is correlated to the pose estimation result, but it is independent of coordinate frames. It is evaluated by comparing predicted distances between endcaps with measured distances from the motion capture system. Smaller distance error indicates a better shape reconstruction in the local frame.

\vspace{-.1in}
\subsection{Baselines and Ablation Studies}\vspace{-.05in}
The first baseline method is naive ICP performed on each rod. Dummy points are added to improve stability because this baseline can easily lose track without them. However, cable length measurements are not used, and no joint optimization is performed. The second baseline first estimates the shape of the tensegrity robot using the endcap distance measurements from on-board sensors. A constrained optimization is done which minimizes the mean squared error (MSE) between the predicted distances between endcaps and the measured ones from on-board sensors, while it constrains the distance between the two endcaps of each rod as the rod length $l_{rod}$. The reconstructed robot is then considered as a rigid body and ICP is used for registration. The third baseline considers all the constraints in section~\ref{physical_constraints}, and performs a correction step similar to the proposed method. The correction step, however, is not integrated in the iterative process. Instead, it first performs ICP for a few iterations and then does one correction step at the end.

In addition, a set of ablation studies are done to verify the effectiveness of the proposed physical constraints and adaptive weights. The first ablation study removes the ground constraint and physical constraints between rods. The second ablation study only removes the physical constraints. The third ablation study keeps all constraints, but uses a static weight (0.25) instead of dynamic weights in the correction step. In addition to those ablation studies, the error of distance between endcaps, which is measured by the elastic stretch sensors, is reported. Note that all the hyperparameters, such as max correspondence distance thresholds, remain the same for all baselines and all testing trajectories for a fair comparison. 

\vspace{-.2in}
\begin{table}[h]
\centering
\begin{tabular}{|c|c|c|c|c|c|}
\hline
    Method & \hspace{0.01in}Trans. Err (cm)\hspace{0.01in} & \hspace{0.01in}Rot. Err (deg) \hspace{0.01in} & \hspace{0.01in} 2cm-5deg (\%) \hspace{0.01in} & \hspace{0.01in} CoM Err (cm) \hspace{0.01in} & \hspace{0.01in} Shape Err (cm)\hspace{0.01in} \\
\hline
    Baseline 1 &  1.56 / 1.89 & 4.55 / 7.96 & 73.1 & 1.07 / 0.76 & 1.30 / 1.81 \\
\hline 
    Baseline 2 &  1.83 / 1.64 & 5.59 / 6.62 & 59.2 & 1.59 / 1.17 & 1.18 / 1.67   \\
\hline                                                            
    Baseline 3 &  1.10 / 0.75 & 3.23 / 2.91 & 79.2 & 0.81 / 0.44 & 0.92 / 0.91 \\
\hline 
    Ablation 1 &  1.22 / 0.96 & 3.81 / 9.28 & 79.2 & 1.01 / 0.60 & 0.89 / 1.00   \\
\hline                                                              
    Ablation 2 &  1.04 / 0.69 & 3.46 / 9.10 & 82.3 & 0.79 / 0.43 & 0.85 / 0.80   \\
\hline                                                              
    Ablation 3 &  1.02 / 0.60 & 2.88 / 2.16 & 85.3 & 0.79 / 0.42 & 0.84 / 0.68   \\
\hline
    Measured   &    N/A       &     N/A     & N/A  &     N/A     & 1.09 / 1.19 \\
\hline                                                            
    Proposed   &  0.99 / 0.58 & 2.84 / 2.12 & 85.5 & 0.77 / 0.44 & 0.80 / 0.64 \\
\hline
\end{tabular}
\vspace{.05in}
\caption{\footnotesize \textbf{Quantitative results of pose tracking and local shape reconstruction.}}
\vspace{-.5in}
\label{table:pose_estimation}
\end{table} 

\vspace{-.1in}
\subsection{Quantitative Results}\vspace{-.05in}
Quantitative results of pose estimation and local shape reconstruction are shown in Table \ref{table:pose_estimation}. The first baseline, naive ICP, has the worst performance in terms of local shape reconstruction, and the second largest tracking error in the global frame. Even though dummy points are added, it still loses tracking in multiple trials and is not able to recover. The second baseline has an even larger tracking error than the naive ICP, which may seem a little surprising. This is because the on-board sensors are often noisy, and the robot being treated as a rigid body after erroneous shape reconstruction further deteriorates the tracking performance. The third baseline performs much better by considering all the physical constraints and performing a correction step. However, ICP and the correction step are performed separately rather than in an interleaving way. Thus, it has an inferior performance than the proposed approach. 

The first two ablation studies shows the benefits of the ground constraints and constraints between rods. Without both constraints, the result of the first ablation is not satisfactory despite being better than the naive ICP. With ground constraints, the translation error in the second ablation is much reduced but the rotation error remains large. The third ablation introduces constraints between rods, and significantly decreases the rotation error. The proposed method performs even better than the third ablation using dynamic weights in the joint optimization. Overall, the proposed method has a mean translation error within 1~cm, mean rotation error within 3~degrees, 85.5\% of the rod pose estimates within 2cm-5deg error range, and mean center of mass error of 0.77 cm for the robot as a whole. It also has the smallest shape reconstruction error in the local frame, which is reflected by the mean distance error of 0.8 cm between endcaps.

\begin{figure}[h]
    \vspace{-0.2in}
    \centering
    \includegraphics[width=\textwidth]{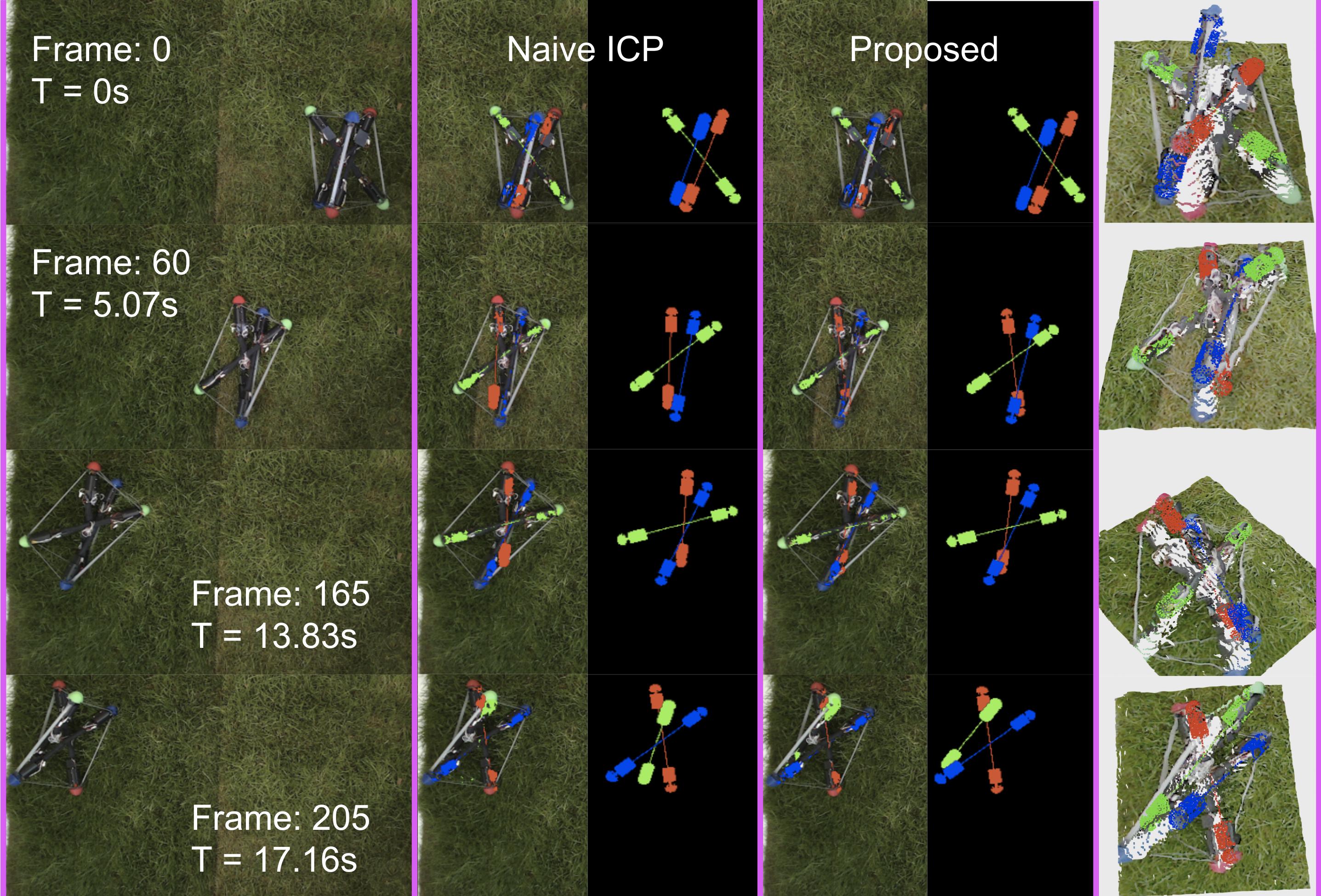}
    \vspace{-.2in}
    \caption{\footnotesize \textbf{Qualitative results.} In this exemplary trial, the tensegrity robot moves from the bottom right to the top left. The left most column shows the raw RGB observation. The second and third columns show the rendered robot at the estimated poses by the naive ICP baseline and the proposed method respectively. The last column show the colored point cloud and robot estimated by the proposed method from some other viewpoints. The proposed method has superior performance over the naive ICP baseline especially when notable occlusion occurs.}
    \label{fig:qualitative_results}
    \vspace{-.3in}
\end{figure}

\vspace{-.1in}
\subsection{Qualitative Results}
\begin{figure}[H]
    \centering
    \includegraphics[width=\textwidth]{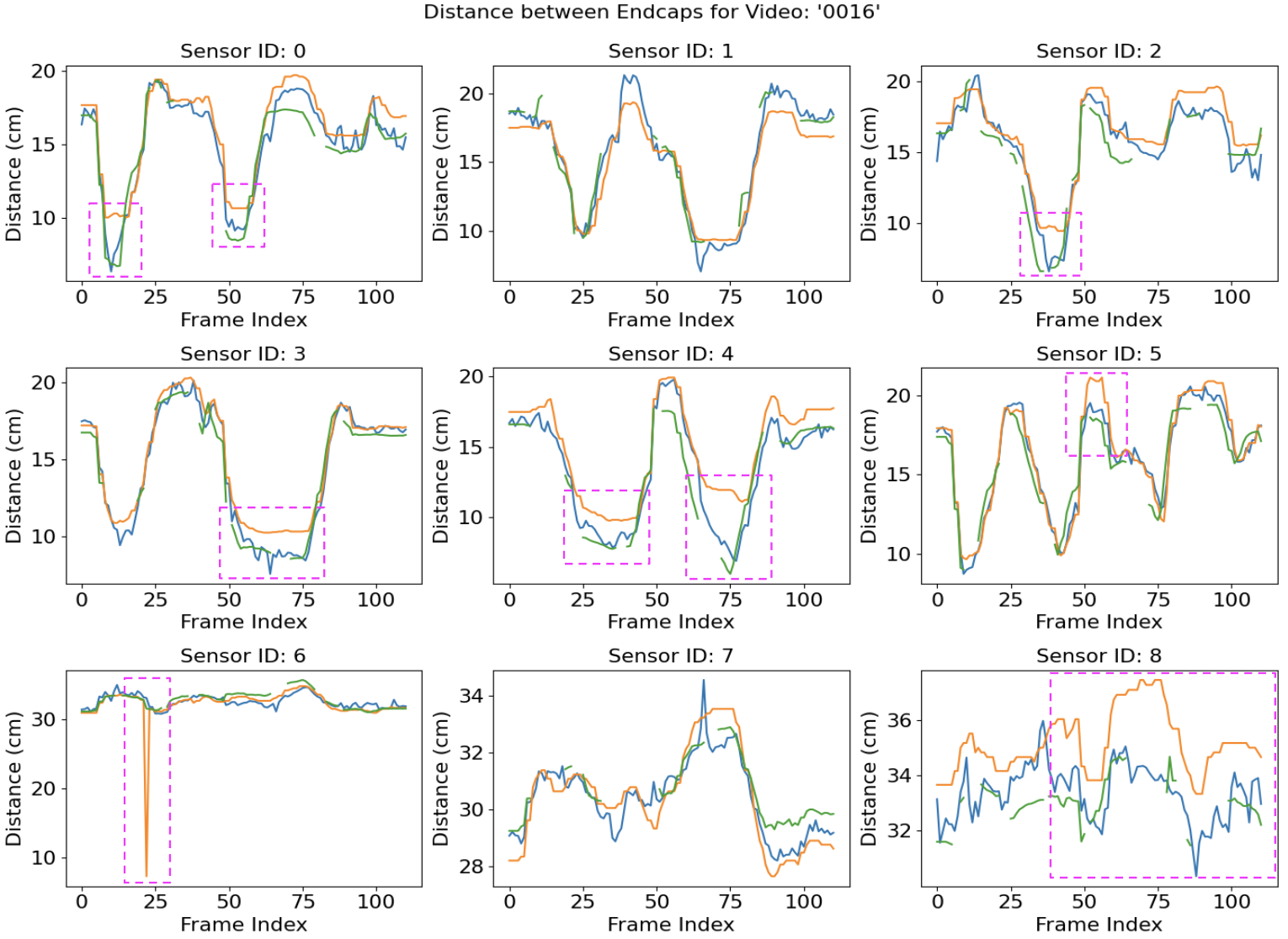}
    \vspace{-.2in}
    \caption{\footnotesize \textbf{Plots of distances between endcaps.} These plots reflect the quality of local shape reconstruction in this exemplary trial. The predicted distances between endcaps by the proposed method (blue lines) are compared with those measured by on-board sensors (yellow lines) and by the motion capture system (green lines, used as ground truth). The predicted distances have even smaller mean error (0.8cm) than that from on-board sensors (1.2cm), which reflects a better shape reconstruction in the local frame. Some notable differences are highlighted in purpose boxes.}
    \label{fig:plot_of_distance_between_endcaps}
    \vspace{-.5in}
\end{figure}

\vspace{-.1in}
An exemplary trial is shown in Fig.~\ref{fig:qualitative_results}, where a robot is moving on a lawn, and the motion capture system is not available. Despite noisy background and self-occlusions, the proposed method is able to give highly accurate pose estimates, while the estimates from the naive ICP baseline have notable errors. In Fig.~\ref{fig:plot_of_distance_between_endcaps}, distances between endcaps at each frame is plotted for another exemplary trial. This is to show how the local shape reconstruction matches with the ground truth. Those predicted distances are compared with the measured distances from the elastic stretch sensors and the motion capture system (ground truth). The predicted results have smaller mean distance error (0.8cm) than those directly measured by the elastic stretch sensors (1.2cm). Some regions are highlighted in purple boxes to show the differences. The regions where the elastic stretch sensors have large errors are typically due to the sensors going slack. The proposed method, however, is robust to those noisy measurements. Fig.~\ref{fig:rod_constraints} shows the effectiveness of physical constraints via another example. Without physical constraint, rods can freely cross through each other. In this example, the red rod is above the green rod and blue rod in the first frame, but it's estimated to be underneath them in the next frame, which is physically infeasible. The blue and green rods are also colliding with each other in this estimate. The proposed physical constraint can effectively prevent such unrealistic estimates from being generated, further improving the accuracy.

\vspace{-.2in}
\subsection{Running Time}\vspace{-.1in}
The running time is computed on a desktop with an AMD Ryzen 5900 CPU. For each iteration, the transition step takes $\sim 2ms$, and the correction step takes $\sim 4ms$. The method can achieve real-time performance ($30Hz$) with 6 iterations at each time step. Note that the accompanying implementation is written in Python, and a faster speed is expected with a more sophisticated implementation in C++.
\pagebreak

\begin{figure}[H]
    \centering
    \includegraphics[width=0.9\textwidth]{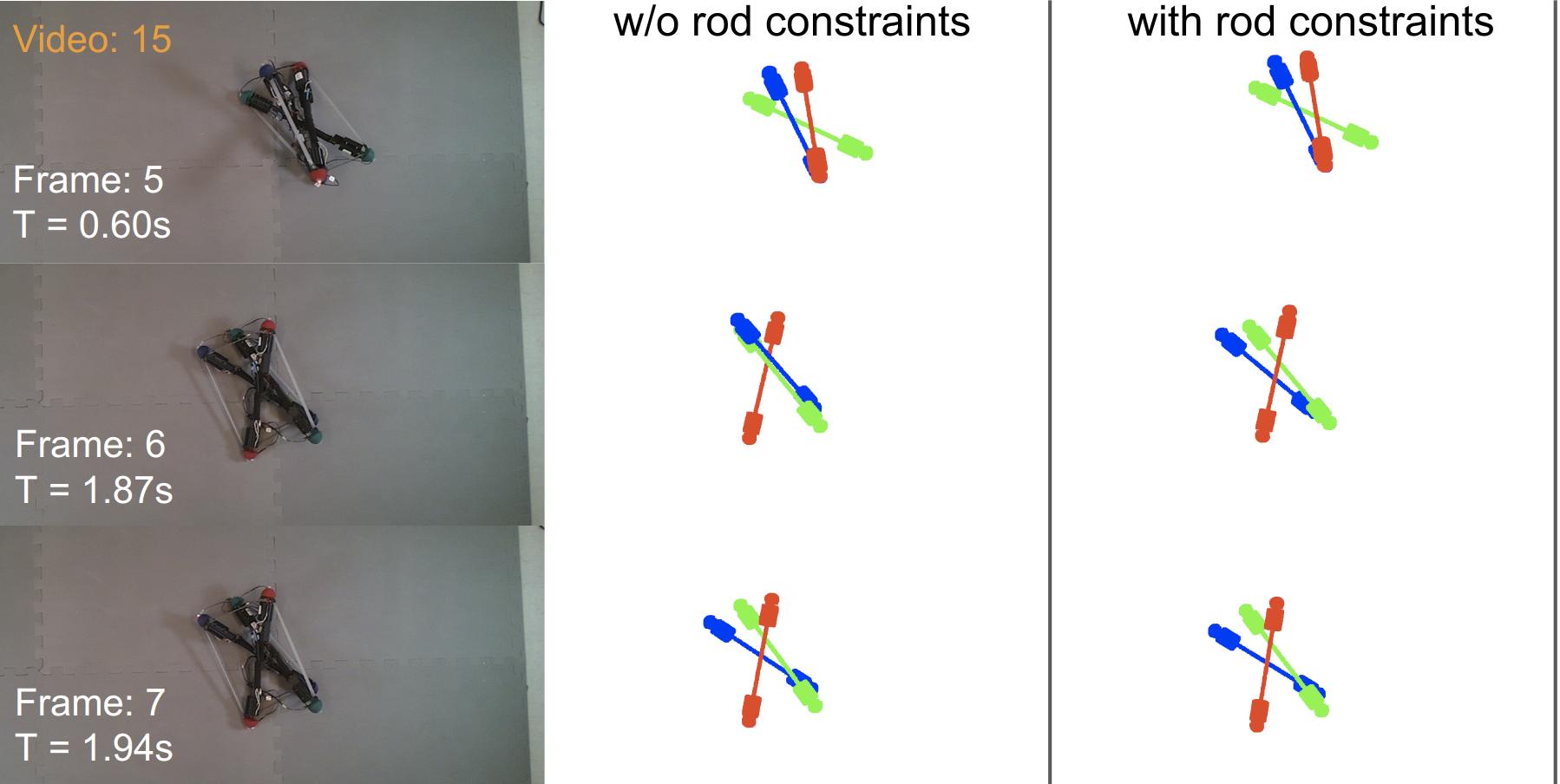}
    \vspace{-.1in}
    \caption{\footnotesize \textbf{Effectiveness of constraints between rods.} This figure compares the pose estimates with and without rod constraints at certain frames. Large error occurs at frame 6 without using rod constraints. The rod with red endcaps crosses through the other two from the previous frame, which is physically infeasible. The blue and green ones are also colliding with each other. As a comparison, the estimates with rod constraints are physical valid and are more accurate.}
    \label{fig:rod_constraints}
    \vspace{-.3in}
\end{figure}
\pagebreak

\section{Limitations}\label{limitations}\vspace{-.1in}

\begin{figure}[h]
    \vspace{-.2in}
    \centering
    \includegraphics[width=0.8\textwidth]{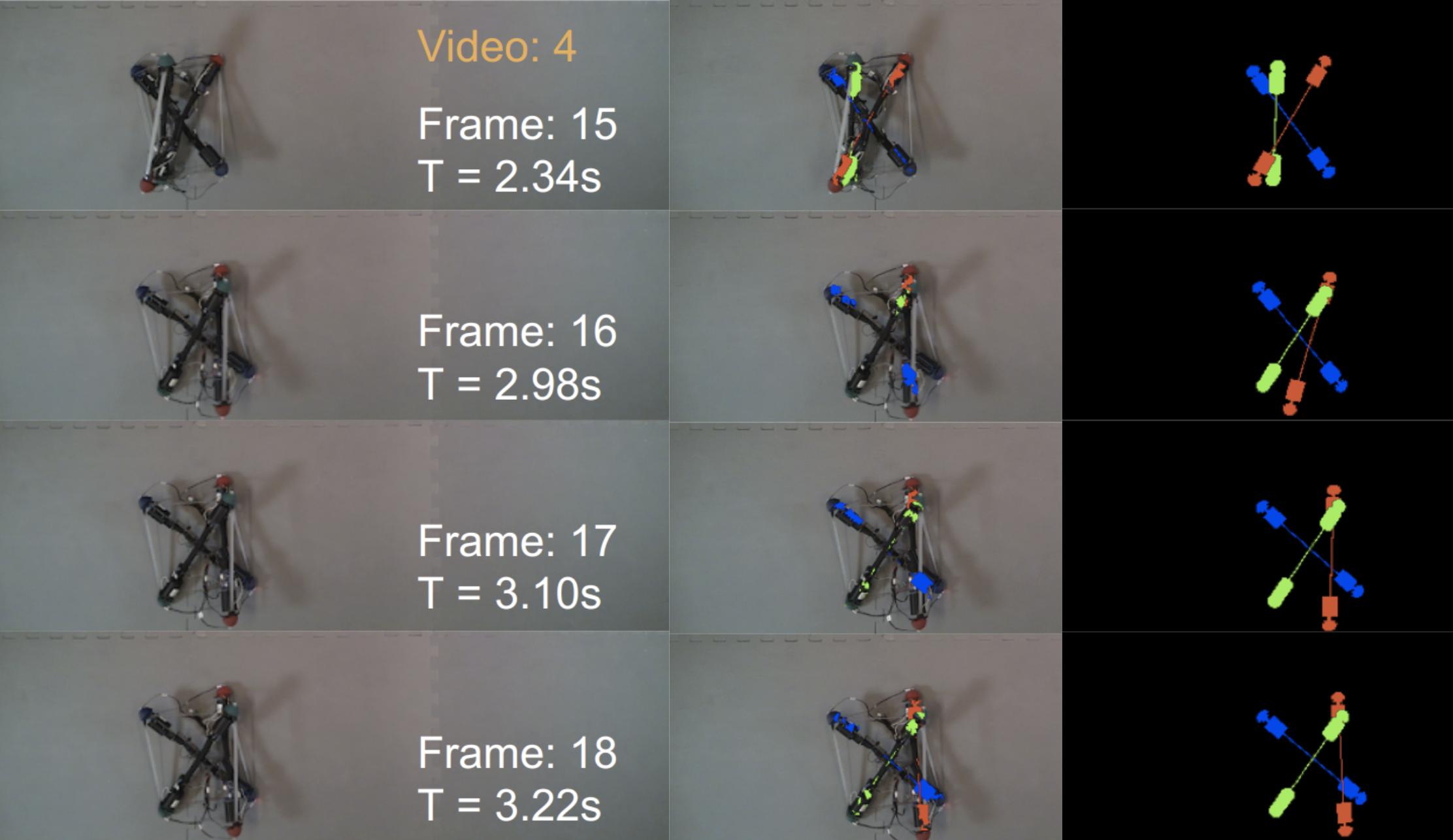}
    \caption{\footnotesize \textbf{Failure case}. A large tracking error occurs at the frame 16 of this video. The robot has a significant movement between two consecutive frames and not enough iterations were performed to make the estimates converge. Tracking is fully recovered later at frame 18.}
    \vspace{-.1in}
    \label{fig:failure_case}
    \vspace{-.4in}
\end{figure}

One limitation of this work is that manual annotation of the 2D region of interest is required for the first frame in order to compute an initial pose. Besides, tracking failure does happen at some time step despite being robust for most of the time. A failure case is shown in Fig.~\ref{fig:failure_case} where the movement of the robot is significant between two consecutive frames. Although the tensegrity robot is not able to move at a high speed, data transmission via Bluetooth can be unstable sometimes. After investigation, it turns out that not enough iterations were performed before the max correspondence distance $d_{max}$ decreases to a value that excludes the observed points. This could be resolved by running more iterations. It is, however, tricky to set such hyperparameters, and there is a trade-off between the number of iterations and the running speed. This work is not data-driven and some manual tuning of these hyperparameters is still required. 

\section{Conclusion}\label{conclusion}\vspace{-.1in}
This work proposes a novel iterative algorithm for 6N-DoF pose tracking of N-bar tensegrity robots. Compared to the standard ICP algorithm for rigid objects, the proposed method fuses the cable length measurement as a soft constraint in the joint optimization process. Meanwhile, a set of physical constraints are introduced to prevent physically invalid pose estimates from being generated. A number of techniques are further applied to improve the robustness of pose estimation against sensor noise. Real-world experiments are performed with an exemplary three-bar tensegrity robot to showcase the robustness and effectiveness of the proposed tracking method, which has a rather small pose estimation error of $1$cm and $3$ degrees for each rod on average. One possible future work is global registration of tensegrity robots, so that the requirement of manual annotation at the initial frame can be removed, and tracking failures can be easily recovered.

%
%
\bibliographystyle{abbrv} 
\bibliography{references} 

\end{document}